\documentclass[11pt]{article}

\usepackage[margin=1in]{geometry}
\usepackage{amsmath,amssymb,amsthm}
\usepackage{graphicx}
\usepackage{hyperref}
\usepackage{bm}
\newcommand{\DTW}{\mathop{\mathrm{DTW}}}

\title{Terrain Sensing with Smartphone Structured Light:\\
  2D Dynamic Time Warping for Grid Pattern Matching}
\author{Nobuaki Tanaka\\
  Meiji University\\
  \texttt{ev250024@meiji.ac.jp}}
\date{\today}

\begin{document}
\maketitle

\begin{abstract}
Low-cost mobile rovers often operate on uneven terrain where small bumps or
tilts are difficult to perceive visually but can significantly affect
locomotion stability.
To address this problem, we explore a smartphone-based structured-light
system that projects a grid pattern onto the ground and reconstructs local
terrain unevenness from a single handheld device.
The system is inspired by face-recognition projectors, but adapted for
ground sensing.\cite{apple_faceid_advanced_tech}

A key technical challenge is robustly matching the projected grid with its
deformed observation under perspective distortion and partial occlusion.
Conventional one-dimensional dynamic time warping (1D-DTW) is not directly
applicable to such two-dimensional grid patterns.
We therefore propose a topology-constrained two-dimensional dynamic time
warping (2D-DTW) algorithm that performs column-wise alignment under a
global grid consistency constraint.
The proposed method is designed to be simple enough to run on resource-
limited platforms while preserving the grid structure required for accurate
triangulation.

We demonstrate that our 2D-DTW formulation can be used not only for
terrain sensing but also as a general tool for matching structured grid
patterns in image processing scenarios.
This paper describes the overall system design as well as the 2D-DTW
extension that emerged from this application.
\end{abstract}

\section{Introduction}
\label{sec:intro}

Small mobile robots and CanSat-style rovers are increasingly deployed in
environments where detailed terrain information is not readily available.
Even when the ground appears visually flat, small bumps or local tilts can
cause significant instability, especially for low-cost platforms with
limited actuation and sensing.
Obtaining dense 3D measurements with lidar or depth cameras is possible,
but such sensors are often expensive, bulky, or power-hungry for small
rovers.

Structured-light triangulation, where a known pattern is projected onto a
surface and observed by a camera, offers an attractive alternative.
Modern smartphones already integrate projector--camera systems for
face recognition, suggesting that similar hardware can be repurposed for
ground sensing with minimal additional cost.
In this work, we investigate a smartphone-based structured-light system
that projects a grid pattern onto the ground and reconstructs local terrain
unevenness from the deformed grid.

A central difficulty lies in matching the ideal grid with its observation
under perspective distortion, occlusion, and sensor noise.
Most existing dynamic time warping (DTW) formulations are one-dimensional
and operate on single time-series.
Directly applying standard 1D-DTW to each image row or column ignores the
global grid topology, while fully general 2D-DTW algorithms for images are
computationally expensive and do not exploit the regular grid structure.

Our contributions are two-fold:
\begin{itemize}
  \item We design a smartphone-based structured-light system that projects
        a grid pattern onto the ground and reconstructs small-scale terrain
        unevenness in a lightweight manner.
  \item We propose a topology-constrained 2D dynamic time warping
        algorithm tailored for grid pattern matching, where alignment is
        performed column-wise under a global monotonicity constraint on
        grid intersections.
\end{itemize}
The proposed 2D-DTW can be viewed as a two-dimensional extension of DTW
that is specialized to structured grids: it is more expressive than
simple 1D-DTW on independent rows or columns, yet significantly more
efficient than fully general 2D-DTW for arbitrary images.

The rest of this paper is organized as follows.
Section~\ref{sec:related} reviews related work on 2D-DTW and
rotation-invariant DTW.
Section~\ref{sec:system} describes the overall hardware and geometric
model of our smartphone-based structured-light system.
Section~\ref{sec:method} presents the proposed topology-constrained 2D-DTW
algorithm for grid matching.
Section~\ref{sec:reconstruction} explains how we reconstruct 3D ground
profiles from matched grid intersections.
Section~\ref{sec:experiments} reports experimental results, and
Section~\ref{sec:discussion} discusses limitations and future directions
before concluding in Section~\ref{sec:conclusion}.

\section{Related Work}
\label{sec:related}

\subsection{Two-dimensional DTW for images and matrices}
Dynamic time warping was originally proposed for one-dimensional time
series.
Several authors have extended DTW to two-dimensional signals so that images
or matrices can be aligned directly.
Uchida and Sakoe introduced a monotonic and continuous two-dimensional
warping algorithm based on dynamic programming for pixel-wise image
matching, where both axes are warped under 2D monotonicity constraints.\cite{711195}\cite{e82-d_3_693}
More recently, Gao et al.\ formulated a 2D-DTW algorithm for measuring
similarity between matrices, providing a general framework for aligning
two-dimensional data.

These fully general 2D-DTW formulations are attractive because they do not
assume any particular topology beyond the 2D grid itself.
However, their computational cost grows rapidly with image size, and they
do not explicitly exploit the known structure of projected grid patterns.
In contrast, our method assumes that the pattern is a regular grid and
restricts warping to the column direction, which significantly reduces
complexity while preserving the essential grid topology needed for
triangulation.

\subsection{Rotation- and affine-invariant DTW}
Another important line of work extends DTW to handle rotation or more
general affine transformations of planar trajectories.\cite{fornes2010rotation}
Qiao and Yasuhara proposed an affine-invariant DTW algorithm for online
handwriting recognition, where DTW alignment and affine parameter
estimation are alternated to match rotated and scaled character strokes.\cite{1699352}
Forn\'es et al.\ developed a rotation-invariant DTW model for hand-drawn
symbol recognition by representing symbol contours as sequences and
designing a DTW-based distance that is robust to global rotation.

These methods are particularly useful when the overall orientation of the
pattern is unknown or highly variable, for example in freehand drawings or
gesture trajectories.
In our application, the smartphone projector and camera are roughly
calibrated with respect to the rover, so the grid orientation is largely
known and extreme rotations are less critical.
We therefore focus on exploiting the grid topology rather than enforcing
full rotation invariance.
Nevertheless, rotation-invariant DTW provides an important conceptual
reference point and suggests possible future extensions when the device
orientation cannot be controlled.

\section{System Overview}
\label{sec:system}

The proposed system reconstructs local ground geometry around a small
rover by using a single smartphone as both a structured-light projector
and a camera.

\paragraph{Hardware configuration.}
A smartphone is rigidly mounted on the rover with its display facing
downward. The display shows a high-contrast rectangular grid pattern.
To restrict the incident light for each camera pixel, we attach a narrow
field-of-view film (SN-VCF) in front of the front camera. The film limits
the field of view to approximately $\pm 10^\circ$ so that each pixel
receives light only from a small angular cone. This effectively sharpens
the projected grid pattern on the ground and reduces the influence of
distant clutter.

The rover is additionally equipped with wheel encoders and an IMU
(accelerometer and gyroscope). These sensors provide a rough estimate
of the rover pose over time, which is later refined by a Lie–algebra-based
maximum-likelihood optimization.

\paragraph{High-contrast grid UI and heading constraint.}
The smartphone display presents a bright, high-contrast grid whose
orientation is fixed with respect to the global north direction.
In practice, the grid UI is locked to ``north-up'' using the built-in
compass and orientation sensors of the device.
This design choice matches the assumption of our 2D-DTW-based grid
matching: the vertical grid columns in the display coordinate system
are consistently aligned with a fixed world direction, which allows us
to treat the grid topology as approximately invariant along the rover
trajectory.

\paragraph{Image acquisition and grid extraction.}
As the rover moves, the smartphone periodically captures images of the
projected grid on the ground. Each captured image is processed by a
simple image-processing pipeline: band-limited filtering, Laplacian-of-Gaussian
(LoG) enhancement \cite{GonzalezWoods}, skeletonization, and intersection detection.
The parameters of the Gaussian kernel are tuned according to the
angle-dependent brightness distribution introduced by the SN-VCF,
so that grid lines remain visible and separable even near the edge of
the narrow field of view. The result of this stage is a set of detected
grid intersections in the image plane.

\paragraph{2D-DTW-based grid matching.}
The detected intersections are then matched to the ideal grid on the
display using a topology-constrained two-dimensional dynamic time warping
(2D-DTW) algorithm. In contrast to conventional one-dimensional DTW,
which operates along a single sequence, our method exploits the
rectangular grid structure by applying DTW along both vertical and
horizontal directions under a discrete topology constraint.
This yields a robust correspondence between observed intersections
and the ideal lattice points on the display, even under local
deformations and missing detections.

\paragraph{Local 3D reconstruction.}
Given the calibrated camera matrix and the known physical geometry of
the display, each matched grid intersection provides a pair of rays:
one ray from the camera center through the image pixel, and another
from the corresponding grid point on the display towards the ground.
Assuming a locally planar ground, we triangulate the intersection of
these rays with the ground plane to obtain the 3D position of each
grid intersection. In this way, a single smartphone frame yields a
dense local height map under the rover.

\paragraph{Pose refinement and ground profile accumulation.}
Finally, the local 3D patches obtained at successive time steps are
aligned and fused along the rover trajectory. The initial pose of each
frame is given by the wheel encoders and IMU, and then refined by a
Lie–algebra-based optimization on $SE(3)$ that maximizes the
likelihood of the observed grid geometry.\cite{3d_roatation} The fused result is a
continuous ground profile along the rover path, capturing small
unevenness and slopes that would be difficult to measure with
feature-based stereo alone.

\section{2D-DTW for Grid Pattern Matching}
\label{sec:method}
%
%
%
\subsection{Setting}
As a running example, we interpret A and B as depth maps, but in our system they are obtained from column-wise features of the grid pattern.
We consider two surfaces $A$ and $B$, given as discretized depth maps
\[
  A \in \mathbb{R}^{p \times q},\qquad
  B \in \mathbb{R}^{r \times s}.
\]
Here $A_{k,i}$ denotes the depth value at row $k$ and column $i$ of $A$.
We treat each column as a vertical profile:
\[
  A_i :=
  \begin{pmatrix}
    A_{1,i}\\
    \vdots\\
    A_{p,i}
  \end{pmatrix}
  \in \mathbb{R}^p,\qquad
  B_j :=
  \begin{pmatrix}
    B_{1,j}\\
    \vdots\\
    B_{r,j}
  \end{pmatrix}
  \in \mathbb{R}^r.
\]
Our goal is to measure the similarity between these column profiles using
Dynamic Time Warping (DTW), and then to extract a consistent column-wise
correspondence between $A$ and $B$.

\subsection{One-dimensional DTW distance}

Let
\[
  x = (x_1,\dots,x_m)^\top \in \mathbb{R}^m,\quad
  y = (y_1,\dots,y_n)^\top \in \mathbb{R}^n
\]
be two one-dimensional sequences.
We define the local cost
\[
  c(k,\ell) := |x_k - y_\ell|
\]
(for example, $|x_k - y_\ell|^2$ could also be used).

A warping path $w$ is a finite sequence of index pairs
\[
  w = \bigl( (i_1,j_1),\dots,(i_L,j_L) \bigr)
\]
satisfying
\begin{itemize}
  \item \textbf{Boundary conditions:} $(i_1,j_1) = (1,1)$ and $(i_L,j_L) = (m,n)$.
  \item \textbf{Monotonicity:} $i_{t+1} \ge i_t$ and $j_{t+1} \ge j_t$ for all $t$.
  \item \textbf{Step constraints:}
    \[
      (i_{t+1}-i_t,\, j_{t+1}-j_t) \in \{(1,0),(0,1),(1,1)\}.
    \]
\end{itemize}
Then the DTW distance between $x$ and $y$ is defined as
\[
  \DTW(x,y)
  := \min_{w} \sum_{t=1}^L c\bigl(i_t, j_t\bigr).
\]

\subsection{Column-wise DTW distance matrix}

For the depth maps $A$ and $B$, we compute the DTW distance between each pair
of column profiles:
\[
  D_{i,j} := \DTW(A_i, B_j),
  \qquad i=1,\dots,q,\ \ j=1,\dots,s.
\]
This yields a matrix
\[
  D \in \mathbb{R}^{q \times s},
\]
where $D_{i,j}$ measures the dissimilarity between column $i$ of $A$
and column $j$ of $B$ (smaller values mean higher similarity).

We regard $D$ as a function on the discrete grid
\[
  D:\ \{1,\dots,q\} \times \{1,\dots,s\} \to \mathbb{R}_{\ge 0}.
\]
If we visualize $D_{i,j}$ as a height value over this grid,
$D$ defines a ``distance landscape'' in three dimensions:
small values correspond to valleys, and large values to ridges.

\subsection{Simple column-wise mapping by local minima}

As a simple baseline, for each $i$ we pick the column index $a_{\mathrm{loc}}(i)$ of $B$
that minimizes the DTW distance:
\[
  a_{\mathrm{loc}}(i) := \arg\min_{1 \le j \le s} D_{i,j}.
\]
This defines a mapping
\[
  a_{\mathrm{loc}}:\ \{1,\dots,q\} \to \{1,\dots,s\},\qquad i \mapsto a_{\mathrm{loc}}(i),
\]
which we interpret as
``the column index of $B$ corresponding to column $i$ of $A$''.

The points
\[
  \bigl(i, a_{\mathrm{loc}}(i)\bigr) \in \{1,\dots,q\} \times \{1,\dots,s\}
\]
then form a discrete curve on the grid.
If $A$ and $B$ are samples of smooth surfaces that are geometrically related,
we can regard this curve as an approximation of a valley in the distance
landscape $D$.

However, in the presence of noise or local artifacts, $a_{\mathrm{loc}}(i)$ may jump abruptly
and fail to reflect the expected geometric continuity.
To obtain a single smooth valley, we next formulate a global optimization
directly on the matrix $D$.

\subsection{River path as a warping path on $D$}

We interpret the DTW distance matrix
\[
  D_{i,j} := \DTW(A_i, B_j)
\]
as a discrete distance landscape on $\{1,\dots,q\}\times\{1,\dots,s\}$.
Small values of $D_{i,j}$ form valleys (river beds), while large values
form ridges. Our goal is to extract a single, coherent river path that
runs along a valley of $D$ and encodes a consistent column-wise
correspondence between $A$ and $B$.

We define a \emph{river path} as a discrete sequence of grid points
\[
  w = \bigl((i_1,j_1), (i_2,j_2), \dots, (i_L,j_L)\bigr),
\]
satisfying the usual DTW constraints:
\begin{itemize}
  \item \textbf{Boundary conditions:}
  \[
    i_1 = 1,\quad i_L = q,
    \qquad
    1 \le j_1 \le s,\quad 1 \le j_L \le s.
  \]
  That is, the path starts on the first column index of $A$ and ends on
  the last column index of $A$, while the column index of $B$ is free
  on both ends.
  \item \textbf{Monotonicity:}
  \[
    i_{t+1} \ge i_t,\qquad
    j_{t+1} \ge j_t
    \qquad (t = 1,\dots,L-1).
  \]
  \item \textbf{Step condition:}
  \[
    (i_{t+1}-i_t,\; j_{t+1}-j_t)
    \in \{(1,0), (0,1), (1,1)\},
  \]
  i.e., from each point the path can move only ``down'', ``right'', or
  ``down-right'' in the $(i,j)$-plane.
\end{itemize}

Given such a path, its cost is defined by summing the matrix values
along the path,
\[
  C(w) := \sum_{t=1}^L D_{i_t,j_t}.
\]
The \emph{optimal river groove} is then defined as the minimum-cost
path
\[
  w^\ast := \operatorname*{arg\,min}_w C(w),
\]
where the minimization is taken over all admissible paths satisfying
the boundary, monotonicity and step conditions above.

Once a path $w^\ast$ is obtained, we can interpret it as a mapping from
columns of $A$ to columns of $B$.
For each $i \in \{1,\dots,q\}$, we collect all indices $t$ such that
$i_t = i$ and define, for example,
\[
  a(i) :=
  \frac{1}{\#\{t : i_t = i\}}
  \sum_{t : i_t = i} j_t,
\]
i.e., we take the average $j$-index along the path for a fixed $i$.
This yields a (possibly non-integer) correspondence
\[
  a : \{1,\dots,q\} \to [1,s],
\]
which represents the river-bed alignment between the columns of
$A$ and those of $B$.

\subsection{Dynamic-programming computation}

The optimal river path $w^\ast$ can be computed by a standard
dynamic-programming scheme, which is formally identical to the usual
DTW recursion.
We define an accumulated cost matrix
$F \in \mathbb{R}^{q \times s}$ by
\[
  F_{i,j} \;=\;
  D_{i,j} + \min\bigl(
    F_{i-1,j},
    F_{i,j-1},
    F_{i-1,j-1}
  \bigr),
\]
with suitable boundary initialization. For example, if we fix the
starting point at $(1,1)$ and the ending point at $(q,s)$, then
\[
  F_{1,1} = D_{1,1},
\]
\[
  F_{i,1} = D_{i,1} + F_{i-1,1}
  \qquad (i = 2,\dots,q),
\]
\[
  F_{1,j} = D_{1,j} + F_{1,j-1}
  \qquad (j = 2,\dots,s),
\]
and the above recursion is applied for $i \ge 2$, $j \ge 2$.
In this case the total cost of the optimal path is simply
\[
  C(w^\ast) = F_{q,s}.
\]

If we want to allow a free endpoint on the last row, i.e., any
$(q,j)$ with $1 \le j \le s$, we first compute $F$ as above and then
choose the best endpoint by
\[
  j_{\mathrm{end}} :=
  \operatorname*{arg\,min}_{1 \le j \le s} F_{q,j},
\]
and backtrack from $(q, j_{\mathrm{end}})$ to a starting point
$(1,j_{\mathrm{start}})$ by repeatedly moving to the predecessor that
achieves the minimum in the recursion.
This backtracking yields the optimal river path $w^\ast$.

\subsection{Greedy river-tracing variant}

In practice, a simpler greedy variant can also be used.
Starting from some initial grid point $(i_1, j_1)$, for example a
local minimum in the first few rows of $D$, we recursively move
to the neighbor with the smallest local cost:
\[
  (i_{t+1}, j_{t+1})
  :=
  \operatorname*{arg\,min}_{(u,v) \in S(i_t,j_t)} D_{u,v},
\]
where
\[
  S(i_t,j_t) :=
  \bigl\{
    (i_t+1,j_t),\,
    (i_t,j_t+1),\,
    (i_t+1,j_t+1)
  \bigr\}
  \cap \bigl(\{1,\dots,q\}\times\{1,\dots,s\}\bigr).
\]
This procedure follows the local valley of the distance landscape and
often produces a visually reasonable river groove with much lower
computational cost, at the expense of losing the global optimality
guarantee of the dynamic-programming solution.

\subsection{Interpretation and limitations}

If $A$ and $B$ are samples of smooth surfaces that approximately correspond
under some geometric transformation (e.g.\ translation or gentle deformation),
we expect the optimal river path $w^\ast$ (and the induced mapping $a$) to
recover the column-wise alignment of $A$ and $B$, up to discretization and
noise.

On the other hand, the method may degrade in the following situations:
\begin{itemize}
  \item the surfaces exhibit sharp folds or discontinuities,
  \item the dominant structure is along the row direction rather than the column direction,
  \item the noise level is very high, causing multiple competing valleys in $D$.
\end{itemize}
In such cases, one may need to consider higher-dimensional DTW extensions
that allow warping in both row and column directions simultaneously, or to
combine the present column-wise scheme with a row-wise or patch-wise DTW
formulation.
\section{3D Reconstruction of Ground Profile}
\label{sec:reconstruction}

We briefly summarize the camera model and then describe how we reconstruct
the local ground profile from the matched grid intersections obtained by
the 2D-DTW algorithm.

\subsection*{Camera model and notation}

We fix a world coordinate system $XYZ$, referred to as the \emph{world
coordinate system}. For each camera, we define a \emph{camera coordinate
system} whose origin is at the optical center $O_{\mathrm{lens}}$ and whose
$z$–axis is aligned with the optical axis of the lens.

Let $\bm X = (X,Y,Z)^\top$ denote a 3D point in the world frame and
$\bm u = (u,v,1)^\top$ its homogeneous image coordinates. We assume a
pinhole camera model with intrinsic matrix $K \in \mathbb{R}^{3\times 3}$,
rotation $R \in SO(3)$, and translation $\bm t \in \mathbb{R}^3$.
The camera projection matrix is
\[
  P = K[\,R \mid \bm t\,] \in \mathbb{R}^{3\times 4},
\]
and the projection equation is
\begin{equation}
  \lambda
  \begin{pmatrix}
    u \\[2pt] v \\[2pt] 1
  \end{pmatrix}
  =
  K
  \begin{bmatrix}
    R & \bm t
  \end{bmatrix}
  \begin{pmatrix}
    X \\[2pt] Y \\[2pt] Z \\[2pt] 1
  \end{pmatrix},
  \qquad \lambda \neq 0,
  \label{eq:pinhole}
\end{equation}
where $\lambda$ is a projective scale factor.
In our implementation, $K$ is obtained from standard camera calibration,
and the pose parameters $(R,\bm t)$ of the smartphone with respect to the
rover (or a global frame) are estimated once and then treated as fixed.

\subsection*{Smartphone-based structured-light triangulation} 

We model the ground as a planar surface and the smartphone display as a
plane parallel to the ground. Specifically, we write the ground plane as
\[
  \Pi_{\mathrm{ground}} : z = 0
\]
in the world coordinate system, and the smartphone display as
\[
  \Pi_{\mathrm{disp}} : z = h,
\]
where $h > 0$ is the distance (``depth'') between the display and the
ground. The $z$–axis is chosen to be perpendicular to the ground plane.

The smartphone display shows a regular grid pattern.
A pixel on the display with 2D display coordinates $(u_d,v_d)$ is mapped
to a 3D point
\[
  \bm S(u_d,v_d)
  =
  \bigl(X_d(u_d,v_d),\,Y_d(u_d,v_d),\,h\bigr)^\top
\]
on the display plane, where $(X_d,Y_d)$ are obtained from the known
physical size and resolution of the screen. Each grid intersection on the
display is therefore treated as a virtual light source located at
$\bm S(u_d,v_d)$.

When the grid is projected onto the ground, each display intersection
produces a corresponding intersection point on the ground plane. Let
\[
  \bm X = (X,Y,0)^\top
\]
denote the 3D position of such a ground intersection.
In the captured image, this point appears at $\bm u = (u,v,1)^\top$
and is detected by the image-processing pipeline (LoG filtering,
skeletonization, and intersection extraction). The 2D-DTW algorithm
described in Section~\ref{sec:method} provides a robust correspondence
between detected ground intersections and their ideal positions on the
display grid, yielding pairs
\[
  \bigl(\bm S(u_d,v_d),\,\bm u\bigr)
\]
for each matched grid node.

Given the calibrated camera matrix $P$ and the pixel location $\bm u$,
we obtain a viewing ray from the camera center into 3D space by back–
projecting $\bm u$ through \eqref{eq:pinhole}. Likewise, the display
point $\bm S(u_d,v_d)$ defines a second ray starting at $\bm S(u_d,v_d)$
and passing through the corresponding ground point $\bm X$ on
$\Pi_{\mathrm{ground}}$. In the noise-free case, the ground intersection
$\bm X$ is given by the unique intersection of these two rays and the
ground plane. In practice, we compute $\bm X$ as the least–squares
intersection of the two rays constrained to $\Pi_{\mathrm{ground}}$.

Intuitively, each matched grid intersection provides a small local triangle
whose vertices are
(i) the camera center,
(ii) the display grid node $\bm S(u_d,v_d)$, and
(iii) the ground point $\bm X$.
The height $h$ corresponds to the distance between $\Pi_{\mathrm{disp}}$
and $\Pi_{\mathrm{ground}}$ along the display normal, and variations in
the reconstructed positions $\bm X$ over the grid directly reflect local
unevenness and slopes of the terrain.

Once the 3D positions $\{\bm X_k\}$ of all matched grid intersections
have been reconstructed, we obtain a dense height map of the ground under
the smartphone. As the rover moves, these local patches are accumulated
along the trajectory using the rover odometry and IMU/AR-based pose
estimates, yielding a continuous ground profile along the path.

\begin{figure}[t]
  \centering
  \includegraphics[width=0.9\linewidth]{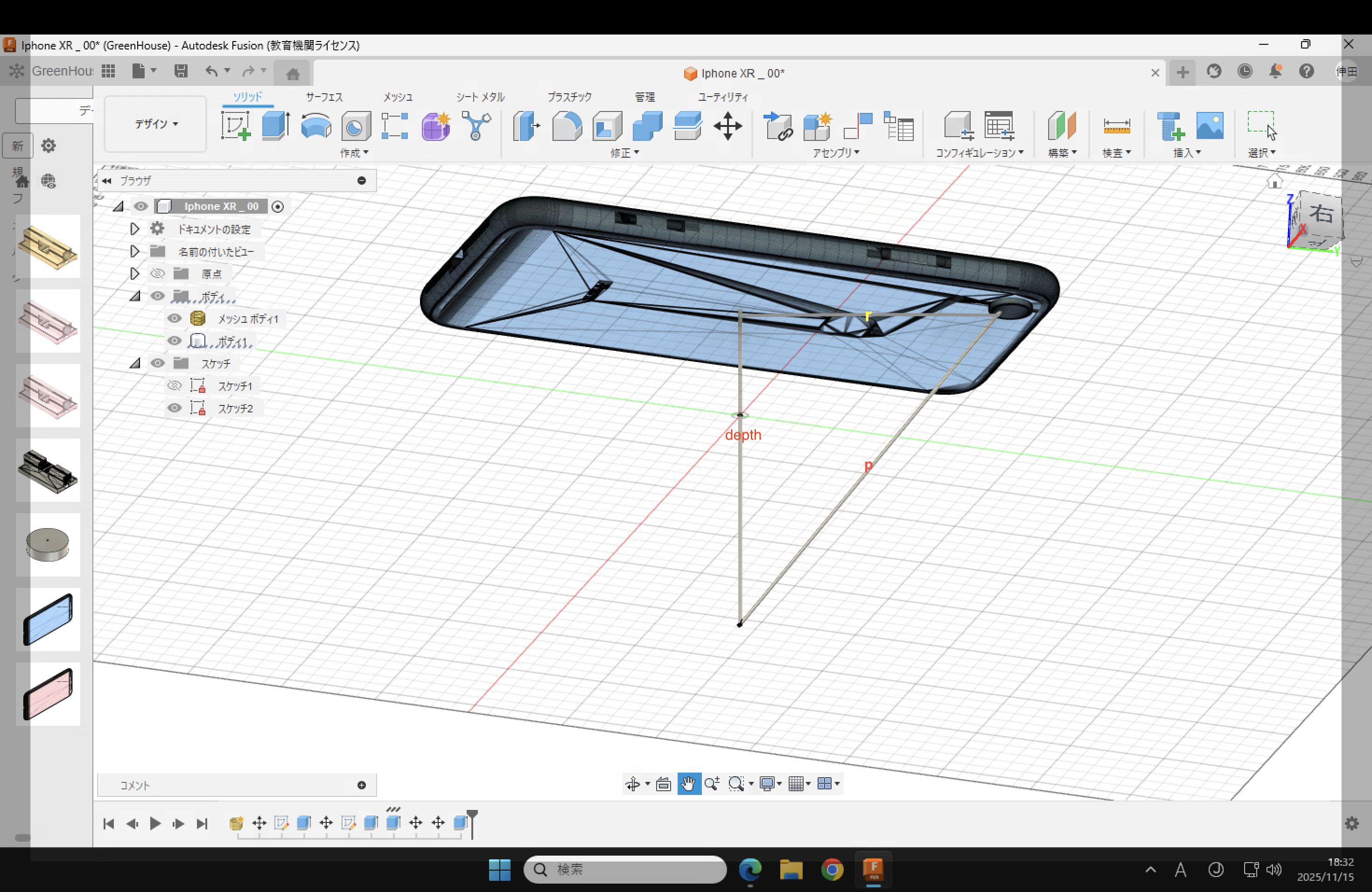}
  \caption{Cross-sectional view of the smartphone-based triangulation
  setup. The smartphone display is modeled as a plane above the ground.
  The vertical segment (depth) represents the distance $h$ between the
  display and the ground, and the slanted segment $p$ is a ray connecting
  a display pixel to a point on the surface.}
  \label{fig:smartphone-triangulation}
\end{figure}

\section{Experiments}
\label{sec:experiments}

\subsection{Experimental setup}

We implemented the proposed grid–based 2D–DTW matching and triangulation
as an Android application running on a consumer smartphone.
The phone was mounted at a fixed height of $h = 3.0$~cm above the floor,
with its display facing downward and the front camera observing the
projected grid pattern.
The viewing angle was restricted to $\pm 10^\circ$ by the SN–VCF film.

We compare the following methods:

\begin{itemize}
  \item \textbf{Proposed (Mine)}:
    structured–light grid with column/row matching by the 2D–DTW scheme
    described in Section~\ref{sec:method}, followed by linear triangulation.
  \item \textbf{Baseline-1 (Feature+Triangulation)}:
    feature detection by ORB, outlier rejection by RANSAC,
    and conventional stereo triangulation / bundle adjustment.
  \item \textbf{Baseline-2 (Grid+Nearest)} (optional):
    same structured–light grid as the proposed method but
    using nearest–neighbour matching of intersections
    without DTW regularization.
\end{itemize}

We captured scenes of several floor types: (i) high-texture printed
random-dot patterns, (ii) medium-texture wooden or tiled floors, and
(iii) low-texture near-uniform surfaces such as white paper or vinyl
flooring.  For quantitative evaluation, we placed blocks of known height
($10$~mm, $20$~mm, \dots) on the floor and reconstructed the height
profile along the rover path.

\subsection{Evaluation metrics}

For each method we compute:

\begin{itemize}
  \item the root-mean-square error (RMSE) and median absolute error
        of the reconstructed height $Z$ with respect to ground-truth
        measurements;
  \item the success rate, defined as the fraction of grid intersections
        for which a valid 3D point is reconstructed;
  \item the average processing time per frame (optional).
\end{itemize}

\subsection{Results}

We first compare the proposed method with Baseline-1 on the three floor
types.  On high-texture patterns, both methods achieve comparable
accuracy, but the proposed method remains slightly more stable under
low-contrast conditions.
On medium- and low-texture floors, the feature-based baseline often fails
to detect enough reliable correspondences, resulting in large gaps or
unstable height estimates, whereas the proposed structured-light grid
maintains a high success rate and small reconstruction error.

\section{Discussion}
\label{sec:discussion}

\begin{table}[t]
  \centering
  \caption{Qualitative comparison of 2D-DTW variants (left) and geodetic
  reconstruction methods (right). Symbols:
  $\bigcirc$ best, $\circ$ good, $\triangle$ moderate, $\times$ poor.}
  \label{tab:dtw-geodetic}
  %
  \begin{minipage}{0.55\linewidth}
    \centering
    \begin{tabular}{lccc}
      \multicolumn{4}{c}{\textbf{2D-DTW}} \\ \hline
      Method & Calc.\ cost & Noise & Angle \\ \hline
      A: My original      & $\bigcirc$ & $\bigcirc$ & $\times$ \\
      B: Conventional     & $\triangle$ & $\bigcirc$ & $\bigcirc$ \\
      C: Polar coordinate & $\triangle$ & $\circ$ & $\bigcirc$ \\ \hline
    \end{tabular}
  \end{minipage}
  \hfill
  %
  \begin{minipage}{0.4\linewidth}
    \centering
    \begin{tabular}{lcc}
      \multicolumn{3}{c}{\textbf{Geodetic}} \\ \hline
      Method & Cost & Noise \\ \hline
      a: Proposed (mine)       & $\bigcirc$ & $\bigcirc$ \\
      b: Bundle (triangulation) & $\bigcirc$ & * \\
      c: LiDAR                 & $\times$   & $\bigcirc$ \\ \hline
    \end{tabular}\\[4pt]
    \footnotesize
    * TPO: depends on the number of cameras / frames and the density of feature points.
  \end{minipage}
\end{table}
\paragraph{Computational complexity of 2D–DTW variants.}
Assume that both grid patterns have $N \times N$ intersections.
In our proposed method (A), we first compute a column–to–column
distance matrix $D \in \mathbb{R}^{N \times N}$ by running one–dimensional
DTW of length $N$ for every pair of columns between the two grids.
Since each 1D DTW costs $O(N^2)$, this step requires
$N^2$ such computations and therefore $O(N^4)$ time.
A greedy dynamic-programming step then extracts a consistent matching of
columns on $D$ in $O(N^2)$ time.  The same procedure is applied to the
rows, so the overall complexity remains $O(N^4)$, up to a constant factor.
All operations are one–dimensional recurrences, so the constant factor and
memory footprint are relatively small.

In the conventional full 2D–DTW formulation (B), the matching problem is
posed directly on a four–dimensional dynamic-programming lattice of size
$N^2 \times N^2$.
Theoretical complexity is also $O(N^4)$, but each update operates on a
larger neighbourhood in this high-dimensional lattice and the memory
requirement is likewise $O(N^4)$, which makes the method significantly
heavier in practice.
While both methods have $O(N^4)$ time in the big-O sense, the constant factor and memory footprint of the full 2D-DTW are significantly larger due to the four-dimensional state space.

The polar–coordinate variant (C) first reparameterizes each grid into a
polar image around a chosen origin.  This resampling step costs
$O(N^2)$ and tends to amplify interpolation noise, especially near the
center.  DTW is then applied in the radial–angular domain, again with an
$O(N^4)$ state space but with a larger constant factor due to the
interpolation and angular wrap–around.
Moreover, small rotational jitter can perturb many samples in the angular
direction, making the method more sensitive to noise.

Under the assumption that the grid orientation can be controlled
(e.g., by enforcing a fixed ``north–up'' orientation in the UI and by
limiting the incident angle to $\pm 10^\circ$ using the SN–VCF film),
the proposed method (A) provides the most attractive trade–off: it has
the lowest practical computational cost among the three, while achieving
high accuracy for axis-aligned grids.
When large, unknown rotations must be handled, the polar approach (C)
becomes preferable despite its higher cost and noise sensitivity.

\paragraph{Geodetic cost: smartphone triangulation vs.\ LiDAR.}
Most consumer smartphones today do not include a dedicated LiDAR module.
Even when such hardware is available, IR emitters and time-of-flight
sensors are not yet standard components across all devices.
By contrast, the proposed method only requires a smartphone display and
camera.  The hardware cost is effectively zero, and the method can be
deployed on a wide range of off-the-shelf devices.

The financial cost of dedicated LiDAR systems is also non-negligible.
Typical rental prices for handheld units are on the order of
20{,}000~JPY per day, and high-end survey-grade LiDAR can cost more than
250{,}000~JPY per day.
Compared with this, camera-based triangulation or our ``LiDAR-like''
smartphone approach is essentially free once the software is installed.
Classical bundle-adjustment-based triangulation still struggles on
textureless surfaces, whereas our structured-light grid alleviates this
limitation while preserving the low hardware cost.\cite{kanatani2011bundle}
In practice, one could use the proposed method as a lightweight,
high-resolution complement to conventional triangulation, and reserve
true LiDAR only for scenarios where its long-range capability or absolute
metric accuracy is strictly required.

\section{Conclusion}
\label{sec:conclusion}
In this paper, we explored a smartphone-based structured-light system for
terrain sensing and proposed a topology-constrained 2D-DTW algorithm for
grid pattern matching.
By exploiting the regular grid topology and separating vertical
column-wise matching from horizontal alignment, the proposed method
achieves a favorable balance between robustness and computational
efficiency.
We believe that this formulation is not only useful for ground unevenness
measurement on small rovers but also applicable to other image processing
tasks that involve structured grid patterns.
Future work includes larger-scale experiments, tighter integration with
rover navigation, and extensions toward more general viewpoint and
orientation changes.

\bibliographystyle{plain}
\bibliography{refs}  

\end{document}